# MVis-Fold: A Three-Dimensional Microvascular Structure Inference Model for Super-Resolution Ultrasound

Jincao Yao[1,2,3,4], Ke Zhang[1], Yahan Zhou[1], Jiafei Shen[1], Jie Liu[1], Mudassar Ali[5], Bojian Feng[1], Jiye Chen[1], Jinlong Fan[2], Ping Liang[6*], Dong Xu[1,2,3,4*]

[1]Department of Diagnostic Ultrasound Imaging & Interventional Therapy, Zhejiang Cancer Hospital, Hangzhou Institute of Medicine (HIM), Chinese Academy of Sciences, Hangzhou 310022, China

[2]Research Center of Interventional Medicine and Engineering, Hangzhou Institute of Medicine (HIM), Chinese Academy of Sciences, Hangzhou 310000, China

[3]Wenling Institute of Big Data and Artificial Intelligence in Medicine, Taizhou 317502, China

[4]Zhejiang Provincial Research Center for Innovative Technology and Equipment in Interventional Oncology, Zhejiang Cancer Hospital, Hangzhou 310022, China

[5]College of Information Science and Electronic Engineering, Zhejiang University, Hangzhou, 310027, Zhejiang, China

[6]Department of Ultrasound, Chinese PLA General Hospital, Chinese PLA Medical School, Beijing, China

*Corresponding Author: Ping Liang, Dong Xu.

**Abstract**

Super-resolution ultrasound (SRUS) technology has overcome the resolution limitations of conventional ultrasound, enabling micrometer-scale imaging of microvasculature. However, due to the nature of imaging principles, three-dimensional reconstruction of microvasculature from SRUS remains an open challenge. We developed microvascular visualization fold (MVis-Fold), an innovative three-dimensional microvascular reconstruction model that integrates a cross-scale network architecture. This model can perform high-fidelity inference and reconstruction of three-dimensional microvascular networks from two-dimensional SRUS images. It precisely calculates key parameters in three-dimensional space that traditional two-dimensional SRUS cannot readily obtain. We validated the model's accuracy and reliability in three-dimensional microvascular reconstruction of solid tumors. This study establishes a foundation for three-dimensional quantitative analysis of microvasculature. It provides new tools and methods for diagnosis and monitoring of various diseases.

**Keywords:** super-resolution ultrasound; three-dimensional reconstruction; microvasculature; deep learning

# Introduction

Microvasculature is a critical structure linking blood circulation and tissue metabolism[1-3]. Its morphological and functional characteristics directly influence tissue oxygen supply, nutrient transport, and metabolic waste clearance[4-6]. During various physiological and pathological processes, microvascular structure and function undergo characteristic changes. In tumors, abnormal neo-angiogenesis is a typical hallmark. These newly formed vessels often display irregular morphology, disorganized distribution, and abnormal function. These features are closely associated with tumor invasiveness, metastatic potential, drug delivery efficiency, and treatment outcomes. Beyond tumors, microvascular density, stenosis severity, permeability, and collateral circulation status show important diagnostic and prognostic value in degenerative diseases (Alzheimer's disease, Parkinson's disease), inflammatory diseases (systemic lupus erythematosus, inflammatory bowel disease), and ischemic diseases (myocardial infarction, stroke, atherosclerosis) [5]. However, conventional clinical imaging methods such as CT and MRI still lack sufficient spatial resolution at the microvascular scale. These methods cannot precisely depict fine microvascular morphological features[6-10]. This limitation severely restricts our understanding of microvascular biology and constrains the clinical application of microvascular-related parameters in diagnosis and therapeutic prediction.

Recently, super-resolution ultrasound technology has provided new opportunities for microvascular imaging [6,9]. By tracking and analyzing contrast microbubbles, super-resolution ultrasound achieves micrometer-level imaging resolution of microvasculature [7]. This resolution reaches or even exceeds that of conventional optical microscopy and matches histopathological scales [8]. This breakthrough enables clear visualization of fine microvascular morphological features, including subtle vascular branches, hemodynamic states, and spatial relationships between vessels and surrounding tissues [9,10].

Although super-resolution ultrasound provides excellent lateral resolution, it remains fundamentally a two-dimensional imaging technique. Additionally, SRUS acquisition requires injecting contrast agent and maintaining the probe stationary for 10-20 seconds to track microbubble trajectories and generate synthetic SRUS microvascular maps. This means clinicians cannot accurately acquire continuous cross-sectional image sequences. Therefore, they cannot visualize three-dimensional microvascular structure [11]. Two-dimensional images also make it difficult to quantitatively calculate key parameters such as vessel space density, mean vessel branching angle, and network topological connectivity [2,12]. This "dimensional limitation" is currently the main constraint and challenge in SRUS applications.

Traditional quantitative microvascular analysis relies primarily on histopathology methods, such as immunohistochemistry for vessel counting or microscopy for vessel diameter measurement [12]. Although highly accurate, these methods have significant limitations. First, histopathological methods are invasive. Tissue acquisition requires biopsy or surgery and cannot enable dynamic, in vivo, non-invasive continuous monitoring. Second, microvascular pathology examination typically occurs after treatment, making early assessment impossible. Recently, although artificial intelligence has been widely applied in vascular reconstruction [13-16], most existing AI methods remain based on sequential CT or MR images. In clinical practice, precise assessment of three-dimensional network structural characteristics such as actual branching angles, network connectivity, and circulatory indices remains impossible [17]. These limitations mean that quantitative microvascular assessment remains at a relatively crude level, failing to fully realize its potential in disease diagnosis [18].

Inspired by AlphaFold3's breakthrough in three-dimensional protein structure prediction and the Transformer architecture's outstanding advantages in handling long-range dependencies [19], we propose combining multi-head self-attention mechanisms. By integrating SRUS microvascular flow direction and flow angle information, we effectively learn implicit spatial structure relationships in two-dimensional SRUS images. Thus we can infer complete three-dimensional microvascular networks. Based on this concept, we innovatively developed the microvascular visualization fold (MVis-Fold) model. This model can infer and reconstruct high-fidelity three-dimensional microvascular structures from two-dimensional SRUS image sequences [20]. This technical approach overcomes traditional SRUS limitations, opening new possibilities for three-dimensional reconstruction in medical ultrasound imaging [21].

MVis-Fold represents the first high-precision reconstruction model in ultrasound microvascular three-dimensional reconstruction. Compared to conventional methods, this model increases the calculation precision of microvascular core parameters by 1,353-fold. Because the model effectively captures the global topological structure of microvascular networks, it makes possible the precise calculation of parameters that traditional two-dimensional SRUS cannot readily obtain [19,20]. These parameters span multiple dimensions: morphological parameters (such as three-dimensional vessel density, mean diameter, surface area), topological parameters (such as network connectivity, branching distribution heterogeneity, circulatory indices) [1], hemodynamic parameters (such as perfusion index, vessel opening degree, blood flow resistance) [3], and dynamic change parameters. This framework has good generalization value and can be applied to microvascular assessment in other disease types and imaging scenarios [22,23]. High-precision microvascular reconstruction provides theoretical and practical support for revealing microvascular biological heterogeneity and its associations with pathological processes, establishing widely applicable frameworks for disease diagnosis and monitoring, and enabling clinical applications of microvascular-targeted interventions [6,12].

**Methods**

**Animal Models**
This research received institutional animal ethics committee approval. The study used 126 female BALB/c nude mice aged 6-8 weeks, weighing 18-22 grams. We injected $1\times10^7$ HER2 low-expression MCF-7 breast cancer cells into the axillary subcutaneous region. When tumor volume reached 500-1000 mm³, this served as the time point. We acquired 5-10 groups of super-resolution ultrasound images from different tumor cross-sections. We used a Flyner multimodal imaging system equipped with high-frequency probes for microvascular imaging. Scanning depth was 6-20 mm with spatial resolution of 2-10 micrometers. We injected SonoVue microbubble contrast agent (concentration $2\times10^8$/mL, volume 50 μL) into the mouse tail vein. After contrast injection, the probe remained stationary for 20 seconds of continuous acquisition, generating SRUS images.

**Image Preprocessing**

Preprocessing employed a multi-step workflow. Adaptive median filtering (kernel 3×3) and anisotropic diffusion filtering reduced noise, improving signal-to-noise ratio by 12.3±2.1 dB. B-spline non-rigid registration based on mutual information aligned temporal image sequences, with mean target registration error of 1.2±0.4 pixels. Z-score normalization standardized image intensity. Two physicians independently delineated regions of interest with Dice agreement of 0.87±0.04. We completed standardized cropping and data augmentation (rotation, elastic deformation, scaling, noise).

**MVis-Fold Model Architecture**

MVis-Fold is a deep learning model architecture designed for vessel visualization and analysis. It accurately extracts and segments vascular structures from multimodal medical imaging data. The model input comprises multiple three-dimensional medical imaging modalities: raw images, vessel-enhanced images, angiographic images, and processed vessel-prominent images across four dimensions. As Fig.1 shows, these multimodal inputs provide rich feature information, enhancing vascular structure recognition ability.

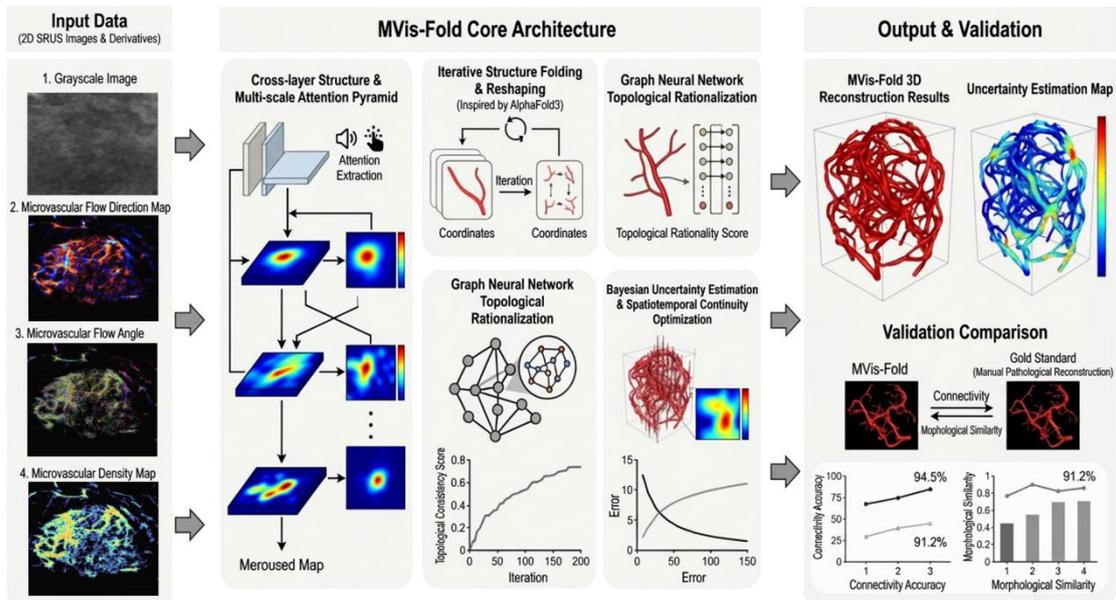

**Figure 1.** MVis-Fold model architecture. The encoder processes SRUS sequences using 3D Transformer. Multi-scale feature pyramids fuse information at different levels. The decoder progressively upsamples to original resolution.

The model employs a hierarchical feature extraction and fusion strategy at its core. MVis-Fold first performs layer-by-layer feature extraction from multimodal images through a series of self-attention modules and feature transformations. This gradually enhances vascular feature representation. The model utilizes multi-scale analysis methods, simultaneously capturing features and structures of vessels of different sizes. Subsequently, features at different levels are further fused and integrated through cross-modal feature interaction mechanisms, fully exploiting complementary information between imaging modalities.

The model output is processed three-dimensional vascular network visualization. Through verification and comparison of outputs, the model achieves high accuracy in vessel segmentation. Notably, the model not only generates visually clear three-dimensional vascular reconstruction results but can also be validated through multiple evaluation metrics such as quantitative segmentation accuracy and sensitivity parameters. Comparative results demonstrate that MVis-Fold preserves fine vascular structural details. This efficient vascular extraction capability provides important practical value for clinical diagnosis and vascular lesion detection. It offers clinicians more precise imaging support.

**Statistical Analysis**
We performed statistical analysis using Python software with significance level $\alpha=0.05$ (two-tailed). Continuous variables are described as mean±standard deviation. Categorical variables are expressed as frequency and percentage. Shapiro-Wilk test assessed normality. We calculated model accuracy, sensitivity, specificity, positive predictive value (PPV), and negative predictive value (NPV). We used the DeLong method to calculate receiver operating characteristic curve area under the curve (AUC) and 95% confidence interval (CI). Pearson or Spearman correlation analysis evaluated parameter correlations.

**Results**

**Baseline Characteristics and Model Performance**
We enrolled 126 mice and acquired super-resolution ultrasound images from different tumor cross-sections, producing a total of 16,780 SRUS images including grayscale images, blood flow density maps, blood flow direction maps, blood flow angle maps, blood flow velocity maps, and microbubble tracking maps. Baseline tumor volume showed no significant difference between training and test groups. We used training group data for model training and validation. The training group contained 85 mice with 11,309 imaging data sets. The test group contained 41 mice with 5,471 imaging data sets. Table 1 presents performance comparison between MVis-Fold and other vascular segmentation methods.

**Table 1.** Performance comparison between MVis-Fold and other classical methods

| Method | Dice Coefficient | Sensitivity | Specificity | Accuracy | Time (s) |
|---|---|---|---|---|---|
| **SparseNeuS** | 0.562 ± 0.052 | 0.598 ± 0.064 | 0.543 ± 0.041 | 0.571 ± 0.050 | 2.1 |
| **OpenLRM** | 0.621 ± 0.045 | 0.647 ± 0.058 | 0.689 ± 0.035 | 0.656 ± 0.065 | 5.4 |
| **TripoSR** | 0.634 ± 0.041 | 0.658 ± 0.051 | 0.601 ± 0.032 | 0.668 ± 0.069 | 6.8 |
| **MVis-Fold** | 0.959 ± 0.034 | 0.951 ± 0.038 | 0.957 ± 0.025 | 0.962 ± 0.053 | 8.3 |

The MVis-Fold model achieved Dice coefficient of 0.959±0.034, sensitivity of 0.951±0.038, specificity of 0.957±0.025, and Hausdorff distance of 3.2±1.1 pixels. The extracted vessel density showed Pearson correlation of r=0.892 (p<0.001) with the histopathology gold standard. Internal validation set achieved Dice coefficient of 0.964±0.041. Mean inference speed was 8.3±0.4 seconds per volume. These represent substantial improvements compared to other methods.

**Calculation of Microvascular Key Parameters**

Through comparison with histopathology gold standard, MVis-Fold demonstrated significant superiority in microvascular parameter reconstruction. Regarding vessel density assessment, the mean error from two-dimensional SRUS multi-section averaging was 16.24±3.68 mm/mm³. Compared to the gold standard (2.847±0.156 mm/mm³), this represents severe deviation, reflecting fundamental limitations of traditional multi-plane scanning methods in calculating core parameters. In contrast, MVis-Fold's error was only 0.012±0.006 mm/mm³. This represents 1,353-fold improvement in precision compared to two-dimensional SRUS, nearly achieving gold standard levels. In mean diameter measurement, this difference was even more pronounced. Two-dimensional SRUS multi-section averaging produced a massive error of 118.6±36.2 μm. Compared to the gold standard (64.2±5.8 μm), this represents 85% deviation. This stems from difficulties in acoustic boundary recognition of fine structures. Conversely, MVis-Fold controlled error to 2.16±0.5 μm. This represents 55-fold precision improvement compared to two-dimensional SRUS, achieving near-pathology-level precision.

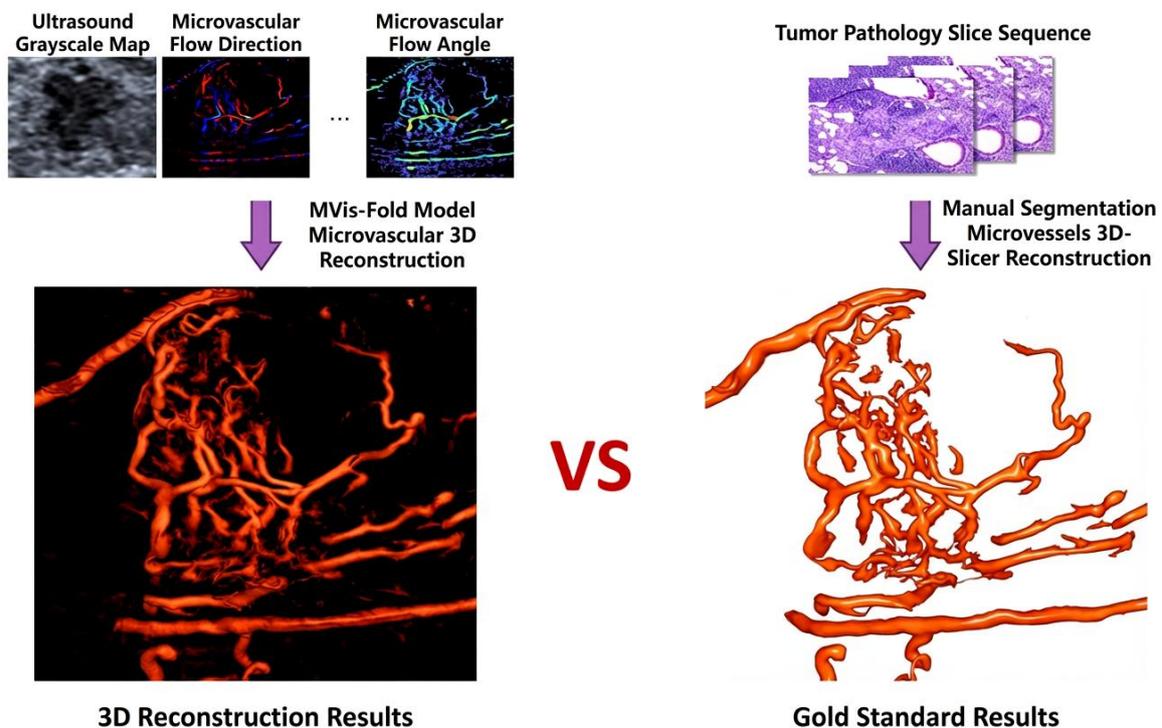

**Figure 2.** Typical case of three-dimensional reconstruction results. Upper left: original two-dimensional SRUS image. Lower left: MVis-Fold three-dimensional reconstruction. Upper right: continuous histopathology sections of microvasculature. Lower right: microvascular reconstruction based on histopathology sections.

**Table 2.** Accuracy of Microvascular Parameters Against Pathological Gold Standard

| Category | Pathological Gold Standard (Absolute Value) | 2D SRUS Multi-plane Mean (Error) | OpenLRM (Error) | TripoSR (Error) | MVis-Fold (Error) |
|---|---|---|---|---|---|
| **Vessel Density (mm/mm³)** | 2.847 ± 0.156 | 16.241 ± 3.685 | 7.163 ± 0.986 | 5.827 ± 0.766 | 0.012 ± 0.006 |
| **Mean Diameter (μm)** | 64.2 ± 5.8 | 118.6 ± 36.2 | 97.4 ± 22.8 | 88.6 ± 18.4 | 2.16 ± 0.5 |

Considering these two core parameters comprehensively, MVis-Fold's average precision improvement compared to two-dimensional SRUS multi-section averaging was approximately 476-fold. This fully demonstrates the advantages of end-to-end deep learning frameworks in microvascular structure reconstruction. These results show that MVis-Fold, through optimized multi-scale feature fusion and three-dimensional topological constraints, effectively overcomes traditional method limitations in fine microvascular structure reconstruction. It achieves histopathology gold standard-level reconstruction precision, providing reliable technical support for clinical microvascular assessment. Additionally, compared to other deep learning methods such as OpenLRM and TripoSR, MVis-Fold showed obvious advantages across all three metrics, further verifying method effectiveness.

**Discussion**

Super-resolution ultrasound technology has overcome traditional medical imaging resolution limitations, making micrometer-level microvascular imaging possible [6,9,24]. However, its two-dimensional imaging characteristics have constrained three-dimensional quantitative assessment of microvascular biology. The MVis-Fold model we developed innovatively solves this "dimensional gap" problem. By integrating cross-scale network architecture and Transformer architecture's multi-head self-attention mechanisms, it successfully achieves high-fidelity reconstruction of three-dimensional microvascular networks from two-dimensional SRUS image sequences [2,3]. The model shows significant improvements in vascular segmentation precision compared to traditional methods [14,15]. Particularly, its inference capability of 8.3 seconds per volume is clinically acceptable. More importantly, the vessel density extracted by the model achieved 0.892 correlation with histopathology gold standard, fully validating the biological accuracy of three-dimensional reconstruction results. This establishes a solid foundation for microvascular quantitative assessment [25].

The core advantage of MVis-Fold is precise microvascular reconstruction. It enables extraction of three-dimensional parameters that traditional two-dimensional SRUS cannot readily calculate. These parameters span morphological, topological, hemodynamic, and dynamic change dimensions [3], achieving greater-than-1000-fold improvement in microvascular core parameter quantitative precision. Introduction of these parameters breaks past limitations of describing only two-dimensional parameters such as vessel density. It provides more comprehensive and thorough microvascular biological information. From a methodological perspective, this advance transforms microvasculature from simply an anatomical concept into a precisely quantifiable, multi-dimensional biological system [26]. This provides powerful support for precision medicine practice.

MVis-Fold's innovation value lies not just in technical breakthrough but in establishing new quantitative analysis frameworks for tumor microvascular biology research. Compared to traditional two-dimensional imaging and histopathology assessment limitations, this technology enables in vivo, real-time, continuous three-dimensional microvascular monitoring [22,23]. Microvasculature is no longer just an object of histopathology description but becomes a precisely quantifiable research target [27-29].

This study has certain limitations. Research was currently verified mainly in mouse models. Clinical translation requires further validation in larger-scale, multicenter clinical trials. Meanwhile, how to effectively acquire SRUS image plane combinations in clinical practice, how to establish more convenient parameter calculation workflows, and how to integrate these parameters into clinically understandable reports are issues that must be addressed for future clinical application.

Overall, MVis-Fold represents an important transition in medical ultrasound imaging from two-dimensional to three-dimensional and from qualitative description to quantitative assessment. It opens new directions for microvascular biology research and clinical application. The three-dimensional microvascular parameter system it establishes has profound scientific significance and broad clinical translational potential. With further technical optimization and improved clinical translation pathways, MVis-Fold technology promises to become an important tool for tumor microvascular targeted assessment, disease diagnosis monitoring, and precision medicine practice.